\documentclass[letterpaper, 10 pt, conference]{ieeeconf}  

\IEEEoverridecommandlockouts                              

\usepackage{amssymb}
\usepackage{amsmath} 
\usepackage{graphicx}
\usepackage{hyperref}
\usepackage{wrapfig}
\usepackage{booktabs}
\usepackage{xcolor}
\usepackage{algorithm}
\usepackage{algpseudocode}

\usepackage[subrefformat=parens]{subcaption}
\usepackage{multirow}
\usepackage{bm}

\usepackage[capitalise,nameinlink]{cleveref}
  \crefname{section}{Sect.}{Sect.}
  \Crefname{section}{Section}{Sections}
  \crefname{figure}{Fig.}{Fig.}
  \Crefname{figure}{Figure}{Figures}
  \Crefname{table}{Table}{Tables}

\title{\LARGE \bf
Refinement of Accelerated Demonstrations via Incremental Iterative Reference Learning Control for Fast Contact-Rich Imitation Learning
}

\author{
    Koki Yamane$^{1, 2}$, Cristian C. Beltran-Hernandez$^{1}$, Steven Oh$^{1, 3}$, Masashi Hamaya$^{1}$, and Sho Sakaino$^{2}$
    \thanks{$^{1}$
        OMRON SINIC X Corporation, 
        Bunkyo-ku, Tokyo 113-0033, Japan 
        {\tt\small masashi.hamaya@sinicx.com}
    }%
    \thanks{$^{2}$
        University of Tsukuba,
        Tsukuba, Ibaraki 305-8573, Japan
    }%
    \thanks{$^{3}$
        Waseda University,
        Shinjuku-ku, Tokyo, 169-8050, Japan
    }%
}

\begin{document}

\maketitle
\thispagestyle{empty}
\pagestyle{empty}

\begin{abstract}
Fast execution of contact-rich manipulation is critical for practical deployment, yet providing fast demonstrations for imitation learning (IL) remains challenging: humans cannot demonstrate at high speed, and naively accelerating demonstrations alters contact dynamics and induces large tracking errors. We present a method to autonomously refine time-accelerated demonstrations by repurposing Iterative Reference Learning Control (IRLC) to iteratively update the reference trajectory from observed tracking errors. However, applying IRLC directly at high speed tends to produce larger early-iteration errors and less stable transients. To address this issue, we propose Incremental Iterative Reference Learning Control (I2RLC), which gradually increases the speed while updating the reference, yielding high-fidelity trajectories. We validate on real-robot whiteboard erasing and peg-in-hole tasks using a teleoperation setup with a compliance-controlled follower and a 3D-printed haptic leader. Both IRLC and I2RLC achieve up to 10$\times$ faster demonstrations with reduced tracking error; moreover, 
I2RLC improves spatial similarity to the original trajectories by 22.5\% on average over IRLC across three tasks and multiple speeds (3$\times$–10$\times$).
We then use the refined trajectories to train IL policies; the resulting policies execute faster than the demonstrations and achieve 100\% success rates in the peg-in-hole task at both seen and unseen positions, with I2RLC-trained policies exhibiting lower contact forces than those trained on IRLC-refined demonstrations.
These results indicate that gradual speed scheduling coupled with reference adaptation provides a practical path to fast, contact-rich IL.
\end{abstract}

\section{Introduction}

\begin{figure}[t]
    \centering
    \includegraphics[width=\linewidth]{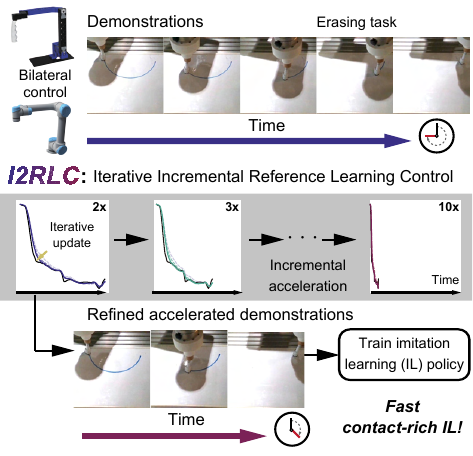}
    \caption{Concept of the proposed method. I2RLC iteratively refines accelerated demonstrations while incrementally increasing the speed. The refined trajectories are then used for fast, contact-rich imitation learning.}
    \label{fig:teaser}
\end{figure}

Imitation learning (IL) acquires manipulation skills from human demonstrations without extensive manual programming~\cite{an2025dexterous}. Recent advances~\cite{chi2023diffusion, zhao2023learning} achieve dexterous manipulation from a few demonstrations by integrating multimodal sensory information, such as vision and contact forces, allowing robots to adapt flexibly to environmental variations. Yet, fast contact-rich manipulation remains challenging: robots must move quickly while keeping interaction forces within safe bounds. Faster execution is critical for practical deployment, as it directly reduces cycle times in industrial applications and enables robots to exceed the speed limits of human demonstrators. This study investigates how to obtain high-speed demonstrations for contact-rich tasks safely, as a step toward deployable, fast contact-rich IL.

A promising strategy for contact-rich IL is to pair a force-controlled follower with a haptic teleoperated leader~\cite{peternel2015human}. Recent studies have developed low-cost systems using bilateral teleoperation with identical manipulators~\cite{yamane2023soft, kobayashi2025alpha, kanai2025input} or an industrial collaborative follower with a 3D-printed haptic leader~\cite{sujit2025improving, liu2025factr}. Whereas prior systems typically rely on torque control, we employ a widely deployed position-controlled arm equipped with a built-in force/torque (F/T) sensor and a compliance controller as the follower, coupled with a 3D-printed haptic leader. This configuration enhances safety; however,  system latency and the demands of precise contact control make it difficult for operators to provide high-speed demonstrations, resulting in datasets dominated by slow-motion trajectories.

One approach is to build a dataset of accelerated demonstrations by downsampling and replaying the control commands at the original control frequency. 
However, such time-scaling alters contact dynamics and often induces large spatial deviations from the original trajectories. A more promising strategy is to execute the accelerated trajectories and refine them using observed tracking errors. Prior methods in this vein either require human intervention~\cite{inami2025motion} or do not address contact-rich tasks~\cite{van2010superhuman}.

This study proposes a method to generate time-accelerated trajectories for contact-rich manipulation without additional human intervention. Our key idea is to \emph{repurpose} Iterative Reference Learning Control (IRLC)~\cite{ducaju2024iterative}, originally developed to reduce tracking error and improve insertion success, to refine accelerated demonstrations. 
However, applying IRLC directly at high speed tends to produce \textbf{larger early-iteration errors}, less stable transients, and \textbf{unsafe behaviors}, risking damage to the environment or hardware and degrading spatial fidelity to the original demonstrations.
To address these issues, we introduce \emph{Incremental Iterative Reference Learning Control (I2RLC)}, which progressively increases execution speed while iteratively updating the reference, shown in~\cref{fig:teaser}. This incremental scheme safely reduces tracking error and yields faster trajectories suitable for contact-rich tasks.

We performed real-robot experiments on several whiteboard erasing and peg-in-hole tasks and compared tracking performance on accelerated demonstrations under IRLC and I2RLC. Both methods refined the reference trajectories and reduced tracking error; however, I2RLC avoided large initial errors and preserved spatial similarity to the demonstrations. We then used the refined trajectories to train IL policies using the method Action Chunking Transformer (ACT)~\cite{zhao2023learning}. 



This study makes the following contributions: 1)IRLC, originally proposed for contact-rich insertion tasks, is repurposed as a general demonstration acceleration framework, with empirical evidence showing that it substantially reduces tracking errors relative to naive speedup playback. 2) I2RLC is introduced as an extension of IRLC that incorporates incremental speed scheduling and warm-start initialization, achieving superior spatial fidelity, particularly at higher acceleration stages, and more stable convergence throughout the iterative refinement process. Both methods are validated on a physical robotic platform across three contact-rich manipulation tasks, demonstrating acceleration factors of up to 10$\times$. I2RLC-refined demonstrations are shown to enable effective training of high-speed imitation learning policies, and a comparison with IRLC-refined demonstrations in the peg-in-hole task shows that I2RLC yields lower contact forces in the trained policies, confirming its benefit for downstream IL applications.

\section{Related Work}
\label{sec:related_work}

\subsection{Fast Contact-Rich Manipulation}
Fast contact-rich manipulation has been achieved via control design, compliant hardware, and simulation-to-real training. A hybrid force–impedance controller with geometry-aware constraints enables torque-controlled robots to execute fast wiping~\cite{iskandar2023hybrid}. High-speed contact manipulation has been demonstrated with highly backdrivable fingers~\cite{karako2019high} and passively compliant arms or grippers~\cite{tanaka2023twist, hartisch2023high, yao2025soft}. Simulation has also been used to train high-speed cutting policies via reinforcement learning~\cite{beltran2024sliceit}. These approaches typically rely on geometric priors, specialized hardware, or simulation. In contrast, this study targets fast contact-rich manipulation on widely deployed position-controlled arms via IL.

Recent IL methods, such as Force-aware ACT variants~\cite{kobayashi2025alpha, liu2025factr, kamijo2024learning} and Diffusion Policy variants~\cite{hou2025adaptive, kang2025robotic}, have demonstrated contact-rich skills including insertion, wiping, pivoting, and prying. Most work emphasizes robustness and generalization; here, we focus on faster execution while maintaining safe contacts. Our approach is complementary: it produces accelerated trajectories that can be used to train these IL backbones.

\subsection{Variable-Speed Imitation Learning}
Several studies address variable speed. Dynamic Movement Primitives expose a time-scaling parameter to generate faster or slower trajectories~\cite{saveriano2023dynamic}. Sakaino {\it et al.} proposed variable-speed IL by conditioning on a speed parameter, showing interpolation and extrapolation to unseen speeds~\cite{sakaino2022imitation}. Extensions adjust the policy's inference frequency to command higher speeds~\cite{inami2025motion}. While these methods are powerful, providing fast demonstrations and extrapolating to substantially higher speeds remains challenging.

\begin{figure}[t]
    \centering
    \includegraphics[width=\linewidth]{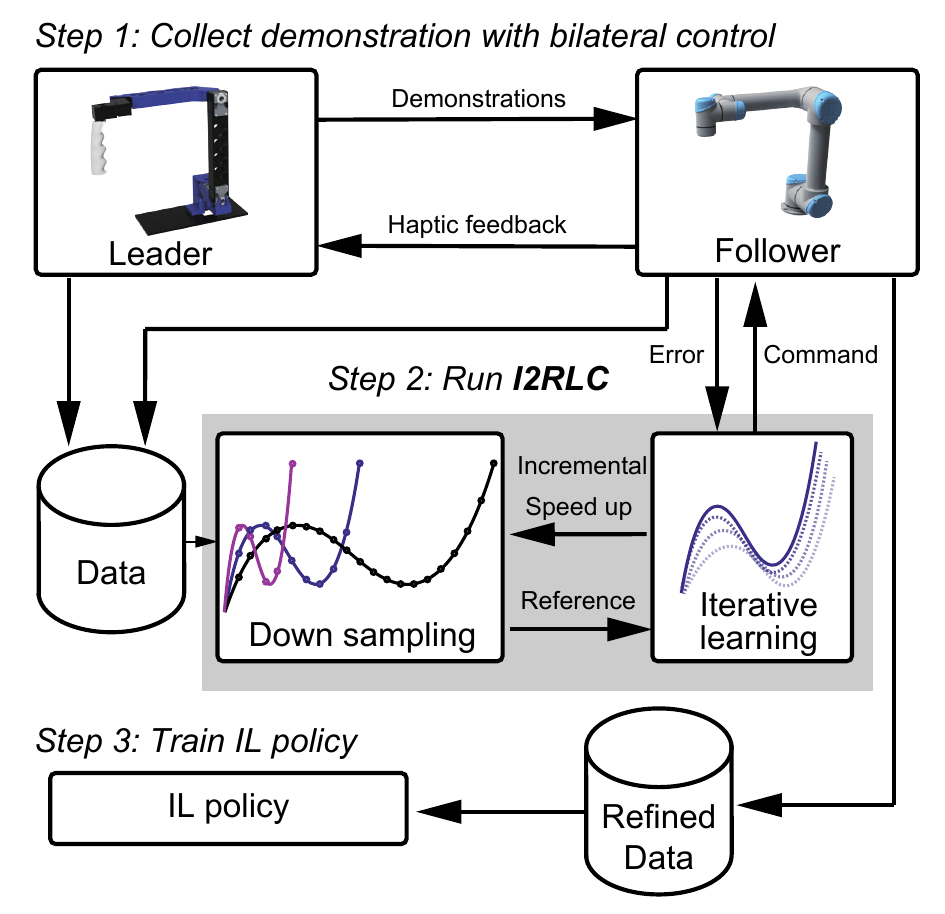}
    \caption{The proposed pipeline comprises three stages: 1) collect demonstrations; 2) downsample to obtain accelerated demonstrations and refine them with I2RLC; and 3) train IL policies on the refined demonstrations.
}
    \label{fig:overview}
\end{figure}

\begin{figure*}[t]
    \centering
    \includegraphics[width=0.98\linewidth]{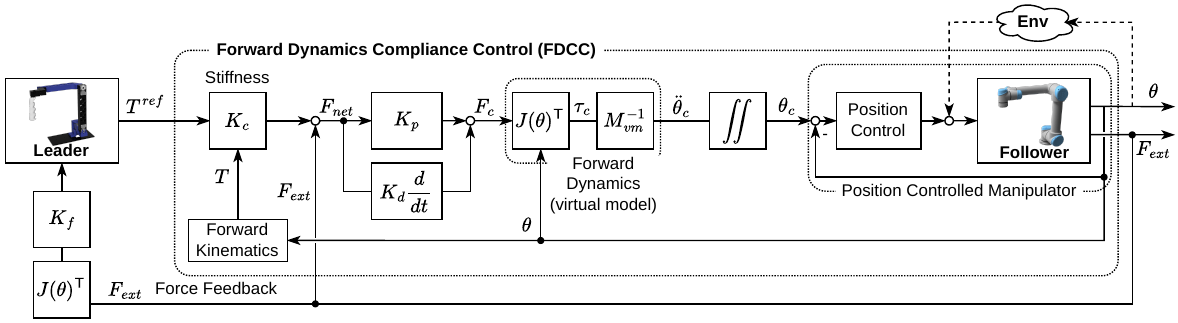}
    \caption{Block diagram of Forward Dynamics Compliance Control (FDCC)~\cite{scherzinger2017forward} and force reflecting type bilateral teleoperation}
    \label{fig:fdcc}
\end{figure*}

Downsampling-based data augmentation can expand coverage~\cite{yamamoto2024real}, but does not guarantee that the augmented trajectories remain executable when contact dynamics change.
DemoSpeedup proposes an entropy-guided acceleration method; however, it focuses on low-precision phases such as free-space approach motions, and thus differs in scope from our study~\cite{guo2025demospeedup}.
SAIL introduces latency-aware scheduling and adaptive speed modulation, achieving up to four times faster execution than demonstrations~\cite{arachchige2025sail}. However, it requires high-gain position control to accurately track accelerated demonstrations, which can generate excessive contact forces and trigger the robot's protective stop.

In contrast, we tackle demonstration acceleration for contact-rich IL that requires no human intervention.

\subsection{Online Trajectory Refinement}
Several approaches can refine demonstrations online. Reinforcement learning can leverage imperfect demonstrations but typically requires substantial data collection~\cite{kang2018policy,ankile2025imitation}. Operator-in-the-loop editing offers another route: \emph{Motion Retouch} overwrites failure segments during time-accelerated execution via multilateral teleoperation~\cite{inami2025motion}, but it relies on human intervention. Iterative Learning Control (ILC) is a classical, data-efficient method for trajectory refinement~\cite{bristow2006survey}. Van den Berg {\it et al.} proposed an LQR-based ILC that gradually speeds up demonstrated trajectories~\cite{van2010superhuman}; however, it assumes linear dynamics and does not account for contact-rich tasks, whose interactions are inherently nonlinear.

We build on IRLC~\cite{ducaju2024iterative} and extend it to I2RLC. IRLC addresses limitations of conventional ILC for contact-rich insertion using an impedance controller. We show that the approach can also refine accelerated trajectories, yielding demonstrations suitable for contact-rich IL.

\section{Methodology}
  Our objective is to generate safe, accelerated demonstrations suitable for contact-rich IL. \cref{fig:overview} illustrates our system's overview. The setup comprises a position-controlled follower, a built-in F/T sensor at the arm’s end effector, a 3D-printed haptic leader, and an RGB camera observing the workspace. An operator first provides demonstrations via the bilateral leader--follower interface. Accelerated trajectories are produced by downsampling the reference motion. Next, IRLC and I2RLC refine these accelerated trajectories. Finally, the refined demonstrations are used for IL.

\subsection{Control System}
  A block diagram of the control system is shown in \cref{fig:fdcc}.

\textbf{3D-printed haptic-feedback leader:}
  A 3D-printed leader providing haptic feedback is employed. The leader builds on the General framework for low-cost and intuitive teleoperation systems (GELLO)~\cite{wu2024gello} and its extensions~\cite{sujit2025improving, liu2025factr}, sharing equivalent kinematics with the follower and utilizing low-cost servomotors with current control. Force-feedback bilateral control is adopted, in which the follower’s measured F/T $\bm{F}_{\mathrm{ext}}$ is mapped to the leader’s joint torques via a feedback gain $\bm{K}_f$. Gravity and friction compensation are also applied to the leader to reduce operator effort.

\textbf{Compliance controller:}
  To enable contact-rich manipulation, Forward Dynamics Compliance Control (FDCC)~\cite{scherzinger2017forward} is employed. FDCC allows a position-controlled robot equipped with an F/T sensor to move compliantly by combining impedance, admittance, and force control. A key component is a virtual forward-dynamics model that maps Cartesian wrench commands to joint accelerations.

  Let $\bm{T}^{ref}$ denote the follower’s reference end-effector Cartesian pose computed from the leader’s joint angles, and let $\bm{K}_c$ denote the Cartesian stiffness. The commanded Cartesian force $\bm{F}_c$ is determined by a desired net wrench $\bm{F}_{\mathrm{net}}$ and PD gains $\bm{K}_{p}, \bm{K}_{d}$. The desired joint acceleration $\bm{\ddot{\theta}}$ is then obtained from the virtual forward-dynamics model given $\bm{F}_c$, the current joint angles $\bm{\theta}$, and a simplified inertia matrix $\bm{M}_{vm}$.

\begin{figure}
    \centering
    \includegraphics[width=0.9\linewidth]{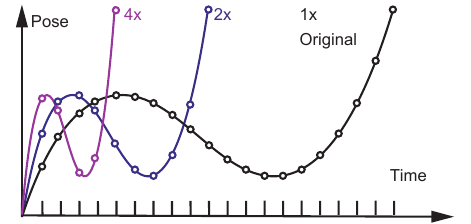}
    \caption{Desired trajectories are downsampled and provided as input to FDCC.}
    \label{fig:downsample}
\end{figure}

\subsection{Data Collection and Demonstration Acceleration}
Demonstrations of length $T$ time steps are collected, forming 
$\mathcal{D}=\{(o_t, \bm{T}_t^{ref})\}_{t=1}^{T}$
, where $o_t$ comprises the F/T readings $F_{ext}$, an RGB image, a homogeneous transformation matrix of the measured follower's Cartesian pose $\bm{T}_t$, and the follower’s joint angles $\theta$. To obtain accelerated demonstrations, the reference trajectory is downsampled by an $n$-fold speedup ($n \in \{1,2, ..., N \}$), resulting in
$\hat{\bm{\tau}}_{n}=\{\bm{T}_{\,nt} \}_{t=1}^{\lfloor T/n \rfloor}$
and
$\hat{\bm{\tau}}_{n}^{ref}=\{\bm{T}^{ref}_{\,nt} \}_{t=1}^{\lfloor T/n \rfloor}$
and
(see \cref{fig:downsample}), where $\lfloor\cdot\rfloor$ denotes the floor function. The FDCC is subsequently commanded with setpoints $\hat{\bm{T}}^{ref}_n$ at a fixed control rate, and $\hat{\bm{T}}_{n}$ is used for IRLC as described in the subsequent section.

\begin{algorithm}[t]
\caption{I2RLC}
\label{alg:i2rlc}
\begin{algorithmic}[1]
\Require Demonstration data ${\mathcal{D}}$; max speed $N$;
         max updates per speed $I$; learning gain $l$
\Ensure Updated reference trajectory $\bar{\bm{\tau}}^{ref}_{(n)}$; measured trajectory $\bm{\tau}$; desired trajectory $\hat{\bm{\tau}}_{(n)}$ (downsampled demo)
\State \textbf{Precompute} accelerated references from the demo:
       $\{\hat{\bm{\tau}}_{(n)},\hat{\bm{\tau}}^{ref}_{(n)} =\textsc{Downsample}(\mathcal{D}, n)\}_{n=2}^N$
\State $\bar{\bm{\tau}}_{(2)}^{ref} \gets \hat{\bm{\tau}}_{(2)}^{ref}$ \Comment{initial reference at $2\times$}
\For{$n = 2$ \textbf{to} $N$} \Comment{incrementally increase speed}
    \If{$n > 2$}
    \State $\bar{\bm{\tau}}^{ref}_{(n)} \gets \hat{\bar{\bm{\tau}}}^{ref}_{(n-1)}$ \Comment{warm start (subsample)}
    \EndIf
  \For{$i = 1$ \textbf{to} $I$} \Comment{updates at $n$-fold speedup}
    \If{$i > 1$}
      \State $\bar{\bm{\tau}}^{ref}_{(n)} \gets \textsc{Update}\big(\bar{\bm{\tau}}^{ref}_{(n)},\,
             \bm{\tau},\, \hat{\bm{\tau}}_{(n)},\, l\big)$
             \Comment{\cref{eq:I2RLC}}
    \EndIf
    \State $\bm{\tau} \gets \textsc{Playback}(\bar{\bm{\tau}}^{ref}_{(n)})$
  \EndFor
\EndFor
\State \Return $\{\bar{\bm{\tau}}_{(n)}\}_{n=2}^N$ \Comment{refined accelerated demo}
\end{algorithmic}
\end{algorithm}

\subsection{Incremental Iterative Reference Learning Control}
Classical iterative learning control suppresses modeling errors and disturbances by augmenting the torque command with a term learned from the previous trial’s tracking error. It is used primarily for high-speed, high-precision position control and has typically been applied to non-contact tasks.

By contrast, recent work proposes Iterative Reference Learning Control (IRLC)~\cite{ducaju2024iterative}, which places an impedance controller at the low level and updates its reference pose rather than directly correcting torques. This design preserves the intrinsic compliance of impedance control while enabling iterative refinement, making it suitable for contact-rich tasks.

The objective is to refine the accelerated demonstrations using IRLC. The update rule can be expressed as follows:



\begin{equation}
\bm{U}^{(i)}_{t} = {^{(i-1)}\bar{\bm{T}}^{ref}_{t}}
\exp\!\left[ l \log \left\{ ^{(i-1)}\bm{T}^{-1}_{t} \hat{\bm{T}}_{nt} \right\} \right]
\end{equation}

\begin{equation}
^{i}\bar{\bm{\tau}}^{ref}_{n} =
\begin{cases}
    \left\{ \bm{U}^{(i)}_{t} \right\}^{\lfloor T/n \rfloor}_{t=1} & (i \geqq 2), \\[4mm]
    \hat{\bm{\tau}}^{ref}_{n} & (i = 1),
\end{cases}
\end{equation}

  where $^{i}\bar{\bm{\tau}}^{ref}$ is the updated reference trajectory at iteration $i \in \{1,2,\ldots,I\}$, then is inputted to FDCC. $l$ is a learning gain. $\exp$ and $\log$ denote the matrix exponential and logarithm, respectively, and these are used to handle 3D rotation.

  This approach is applied to replay recorded demonstrations at high speed. Specifically, the demonstration is downsampled to obtain an accelerated reference, which is then executed while the tracking error relative to the original is measured and used to update the reference. This procedure is repeated over several iterations. However, larger $n$-fold speedups can cause the uncorrected motion to deviate substantially, especially in the early iterations, leading to serious failures in contact-rich tasks. 

  To address this, \emph{Incremental Iterative Reference Learning Control (I2RLC)} is proposed, which gradually increases speed across iterations while updating the reference from observed tracking errors. This incremental scheme enables iterative playback that remains close to the original trajectory even under substantial speedups. The update rule is:





\begin{equation}
\label{eq:I2RLC}
^{i}\bar{\bm{\tau}}^{ref}_{n} =
\begin{cases}
    \left\{ \bm{U}^{(i)}_{t} \right\}^{\lfloor T/n \rfloor}_{t=1} & (i \geqq 2), \\[4mm]
    {^{I}\hat{\bar{\bm{\tau}}}^{ref}_{n-1}} & (i = 1),
\end{cases}
\end{equation}



If $i \geq 2$, the update rule is identical to IRLC. For $i=1$, the reference is warm-started with ${^{I}\hat{\bar{\bm{\tau}}}^{ref}_{n-1}}$, the downsampled reference obtained after $I$ iterations at the $(n-1)$-fold speed stage by uniformly subsampling to match the trajectory length at stage $n$. After completing the iterations at stage $n$, the process advances to the next stage by setting $n \leftarrow n+1$. This warm start mitigates large tracking errors at the initial iteration. \cref{alg:i2rlc} summarizes I2RLC's procedures.


\subsection{IL Training}
   Given the refined accelerated demonstrations, imitation learning policies are trained. Although I2RLC is compatible with arbitrary IL algorithms, this study adopts ACT~\cite{zhao2023learning} because of its sample efficiency. ACT employs a Transformer architecture to predict sequences of future actions, which are stored and temporally ensembled to produce smoother action generation. The ACT input comprises $F_{ext}$, an RGB image, the follower's pose $\bm{T}$, and the output is the sequence of $\bm{T}^{ref}$. A vector representation is used for the 6D pose, consisting of a 3D translation vector and a 3D rotation vector.

\begin{figure}
    \centering
    \includegraphics[width=\linewidth]{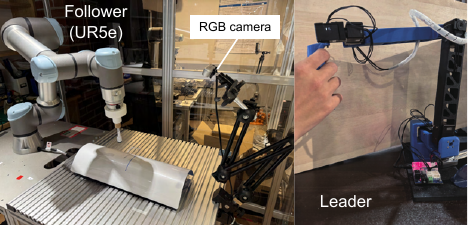}
    \caption{Experimental setup.}
    \label{fig:setup}
\end{figure}
\begin{table}
    \centering
    \caption{Parameters for FDCC controller}
    \label{table:controller parameters}
    \begin{tabular}{c|cccccc}
        \toprule
         & x & y & z & rx & ry & rz \\
        \midrule
        $K_c$ & 300 & 300 & 300 & 100 & 100 & 100 \\
        $K_p$ & 0.0252 & 0.0252 & 0.0252 & 0.36 & 0.36 & 0.36 \\
        $K_d$ & 0.0072 & 0.0072 & 0.0072 & 0.0072 & 0.0072 & 0.0072 \\
        \bottomrule
    \end{tabular}
\end{table}

\begin{table*}[t!]
    \centering
    \caption{Playback, IRLC and I2RLC Trajectory error comparison}
    \label{table:error_comparison}
    \begin{tabular}{ccc|cccccccc}
        \toprule
        && Speed & 3$\times$ & 4$\times$ & 5$\times$ & 6$\times$ & 7$\times$ & 8$\times$ & 9$\times$ & 10$\times$ \\
        \midrule
        \multirow{3}{*}{Flat-surface erasing} 
             &       & Baseline (Playback)   & 4.319 & 4.918 & 5.128 & 5.359 & 5.432 & 5.496 & 5.603 & 5.545 \\
             & DTW  $\downarrow$    & Ours (IRLC)    & 1.064 & 1.251 & 1.007 & 0.855 & 0.697 & 0.652 & 0.612 & 0.615 \\
             &          & Ours (I2RLC)   & \bf{0.585} & \bf{0.511} & \bf{0.479} & \bf{0.536} & \bf{0.483} & \bf{0.539} & \bf{0.447} & \bf{0.419} \\
        \midrule
        \multirow{3}{*}{Curved-surface erasing} 
              &      & Baseline (Playback)   & 5.599 & 5.666 & 5.355 & 6.755 & 5.762 & 6.016 & 6.624 & 6.005 \\
              & DTW  $\downarrow$    & Ours (IRLC)   & \bf{2.103} & 1.765 & 1.460 & 1.765 & 1.239 & 1.092 & 1.193 & 1.035 \\
              &         & Ours (I2RLC)   & 2.157 & \bf{1.733} & \bf{1.423} & \bf{1.140} & \bf{1.012} & \bf{0.923} & \bf{0.785} & \bf{0.817} \\
        \midrule
        \multirow{3}{*}{Peg-in-hole} 
              &      & Baseline (Playback)  & 1.000 & 0.878 & 0.784 & 0.725 & 0.689 & 0.672 & 0.663 & 0.634 \\
              & DTW  $\downarrow$    & Ours (IRLC)   & \bf{0.253} & {0.234} & 0.241 & \bf{0.154} & {0.176} & 0.179 & 0.242 & 0.156 \\
              &         & Ours (I2RLC)   & 0.283 & \bf{0.225} & \bf{0.210} & 0.199 & \bf{0.167} & \bf{0.110} & \bf{0.127} & \bf{0.085} \\
        \bottomrule
    \end{tabular}
    \vspace{-1mm}
\end{table*}
\section{Experiments}
   The evaluation addresses two questions: 1) whether IRLC and I2RLC can effectively refine accelerated demonstrations, and 2) whether the refined trajectories can be used for IL. To this end, real-robot experiments are conducted on contact-rich erasing and peg-in-hole tasks. Tracking performance is compared in \ref{sec:I2RLC_exp}, followed by the training of IL policies using the refined demonstrations in \ref{sec:IL_exp}.

\subsection{Robot System}
   The experimental setup is illustrated in \cref{fig:setup}. A Universal Robots UR5e with a built-in F/T sensor is used as the follower, and an Intel RealSense D435 camera is positioned at a fixed viewpoint. An eraser holder attached to the end effector incorporates springs to promote stable, compliant contact. The leader employed six servomotors: three Dynamixel XM430-W350-T units for the shoulder and elbow joints, and three Dynamixel XC330-T288-T units for the wrist joints.
   The control frequency of the FDCC and the leader was 500 Hz, and the communication frequency between the leader and the follower was 50 Hz. The stiffness parameter and PD gain for FDCC in~\cref{fig:fdcc} are shown in \cref{table:controller parameters}. The control system was implemented on ROS1 Noetic.
   A desktop PC powered by an AMD Ryzen Threadripper 7960X CPU and two NVIDIA RTX 4000 Ada GPUs was used for robot control and training of the IL policies.

\subsection{Tasks}

\textbf{Flat-Surface Arc Erasing:}
A contact-rich, trajectory-sensitive task is considered, in which the robot erases an arc drawn on a whiteboard using an end-effector-mounted eraser. As the target trajectory follows an arc rather than a straight line, naive time-downsampled replay results in significant end-effector path deviation from the original demonstration, potentially leading to task failure.

\textbf{Curved-Surface Arc Erasing:}
As a more challenging contact-rich task, the robot traces and erases an arc on a convex cylindrical surface. High-speed execution is difficult due to the increased likelihood of the eraser lifting off or applying excessive contact force. 

\textbf{Peg-in-hole:}
A peg-in-hole task is considered using a circular peg with a 2 mm tolerance, where high-speed execution tends to amplify contact forces and increase the risk of insertion failure.

\subsection{Incremental Iterative Reference Learning Control}
\label{sec:I2RLC_exp}

\textbf{Setup:}
   IRLC operates at a fixed target speed, whereas I2RLC incrementally increases the speed up to 10$\times$.
   At each speed, three update iterations ($I=3$) are performed with a gain $l=0.4$, which is selected to balance tracking improvement and motion stability. For a fair comparison, the total number of update iterations is matched across both methods. In I2RLC, the refined trajectory obtained at each speed stage is carried forward as the initialization for the subsequent stage.

\begin{figure}[t!]
    \vspace{1mm}
    \centering
    \includegraphics[width=0.95\linewidth]{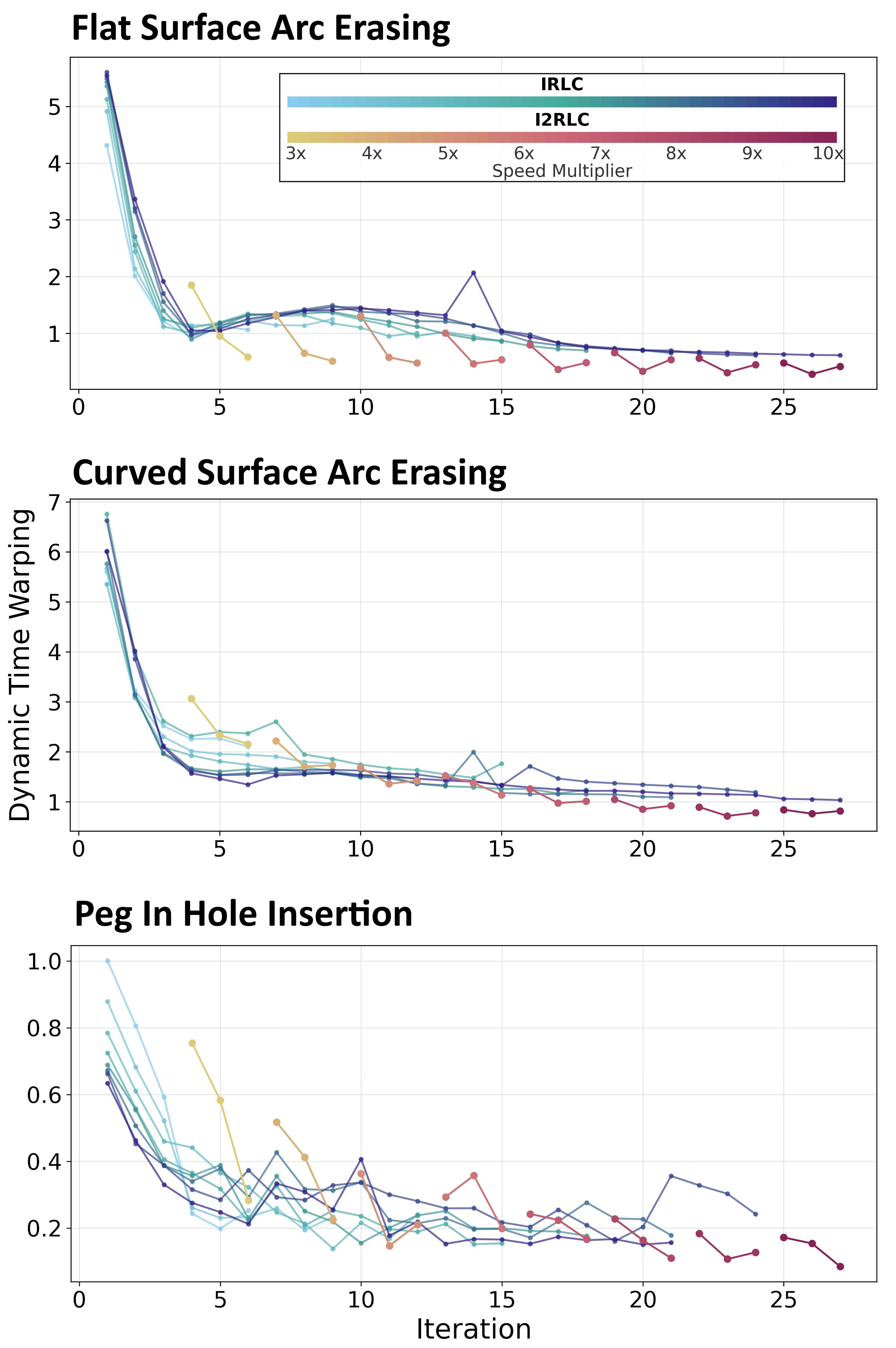}
    \caption{Learning curves based on dynamic time warping for IRLC and I2RLC on the arc-erasing and peg-in-hole tasks. For fairness, the total number of update iterations is matched across methods. I2RLC shows consistently lower DTW.}
    \label{fig:error_comparison_flat_surface}
\end{figure}
\begin{figure*}[t]
    \centering
    \includegraphics[width=0.95\linewidth]{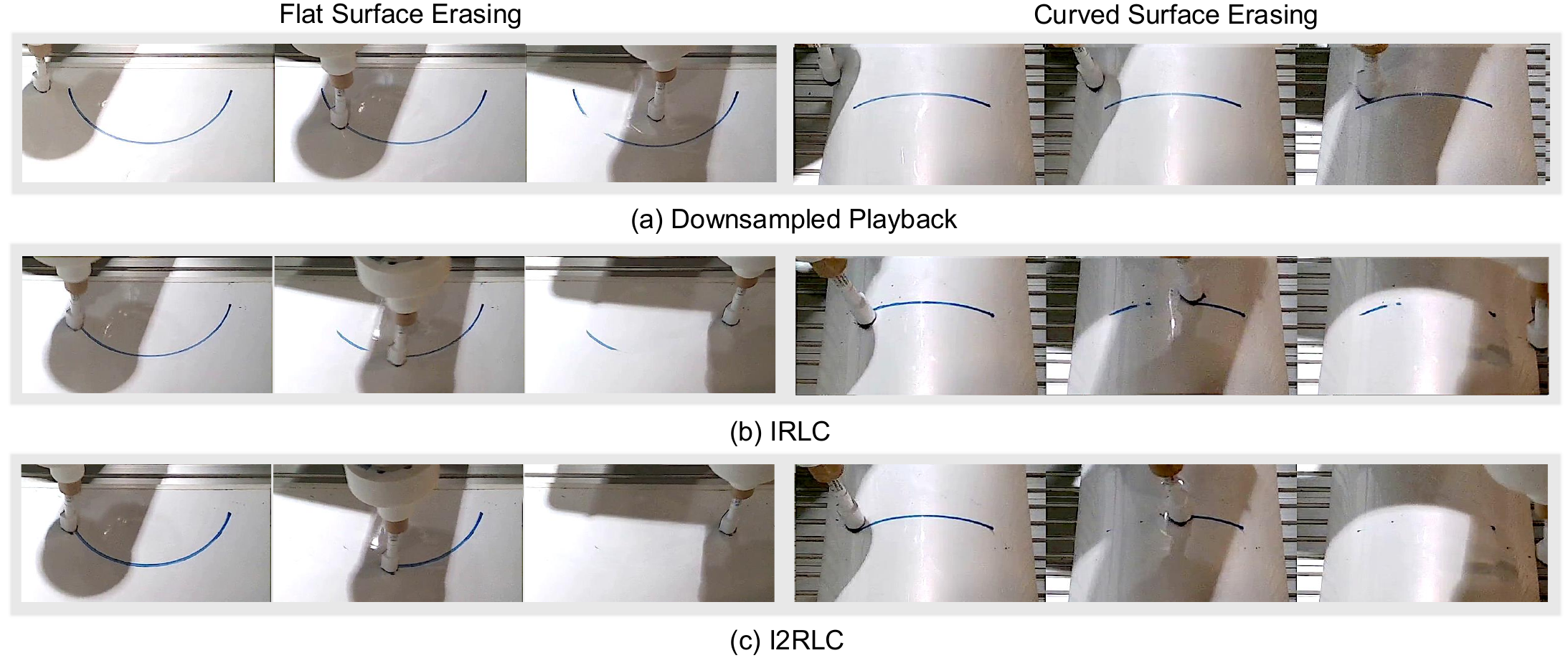}
    \caption{Snapshots of erasing tasks in flat and curved surfaces (10$\times$ speed). I2RLC successfully erased the blue lines.}
    \label{fig:snapshots}
    \vspace{-2mm}
\end{figure*}

\textbf{Metric and baseline:}
   The dynamic time warping (DTW) error of the X, Y, and Z positions is reported as a tracking metric. DTW quantifies spatial discrepancy between trajectories via nonuniform temporal alignment, making it well-suited for assessing demonstration quality. Larger discrepancies can introduce task failures or excessive contact forces.

   Downsampled playback, as employed in~\cite{yamamoto2024real}, is selected as the baseline. As discussed in Sec.~\ref{sec:related_work}, the existing methods~\cite{guo2025demospeedup, arachchige2025sail} are not applicable to our setting, where the robot maintains continuous contact with the environment; in our teleoperation trials with high-gain position control, the robot immediately triggered its protective stop upon contact.

\begin{figure}[t!]
    \centering
    \includegraphics[width=\linewidth]{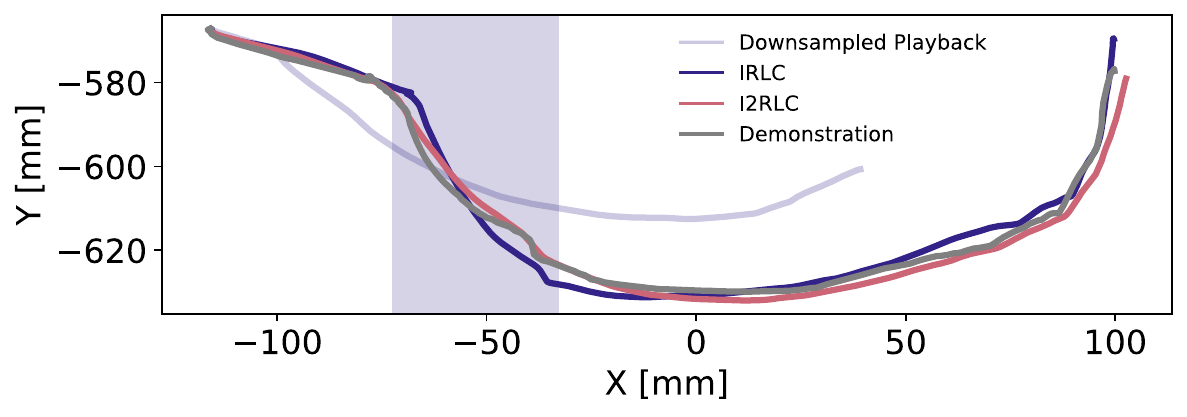}
    \caption{Trajectory comparison for flat surface arc erasing at 10$\times$ speed. Shaded regions indicate segments where IRLC incurs larger errors, corresponding to the residual line segments in \cref{fig:snapshots}(b).}
    \label{fig:trajectory_comparison_falt_surface}
\end{figure}

\textbf{Results:}
For quantitative analysis, 
\cref{table:error_comparison} shows the DTW errors between executed and downsampled reference trajectories at the final iteration of each speed stage. 
We report one run per condition since the refinement update is deterministic under fixed demonstrations and controller parameters. Most experiments achieve up to 10$\times$ speedup. An exception occurs in the peg-in-hole task, where IRLC triggers the robot's protective stop at the 22nd iteration under 10$\times$ speedup; therefore, we report the result up to the 21st iteration at this speed.
Across all tasks, the naive playback baseline yields the largest DTW error. In the erasing tasks, I2RLC achieves the lowest DTW error, outperforming IRLC and indicating better spatial trajectory alignment. 
In the peg-in-hole task, the two methods show mixed results at lower speeds, while I2RLC consistently outperforms IRLC at higher speeds (7$\times$–10$\times$).

\cref{fig:error_comparison_flat_surface} shows the learning curves for IRLC and I2RLC. Both methods reduce DTW error over iterations. Under IRLC, trajectories often deviate significantly in early iterations, producing large errors. By contrast, I2RLC maintains errors within a narrower range throughout, substantially reducing the risk of unstable behavior before convergence. Interestingly, DTW errors under I2RLC decrease at higher speeds, as each iteration is initialized from the accelerated trajectory of the previous one, effectively serving as a warm start. Note that in the peg-in-hole task, DTW errors in early iterations are larger at lower speeds, as the peg tended to remain near the hole for longer durations.

\begin{figure}
    \centering
    \includegraphics[width=0.95\linewidth]{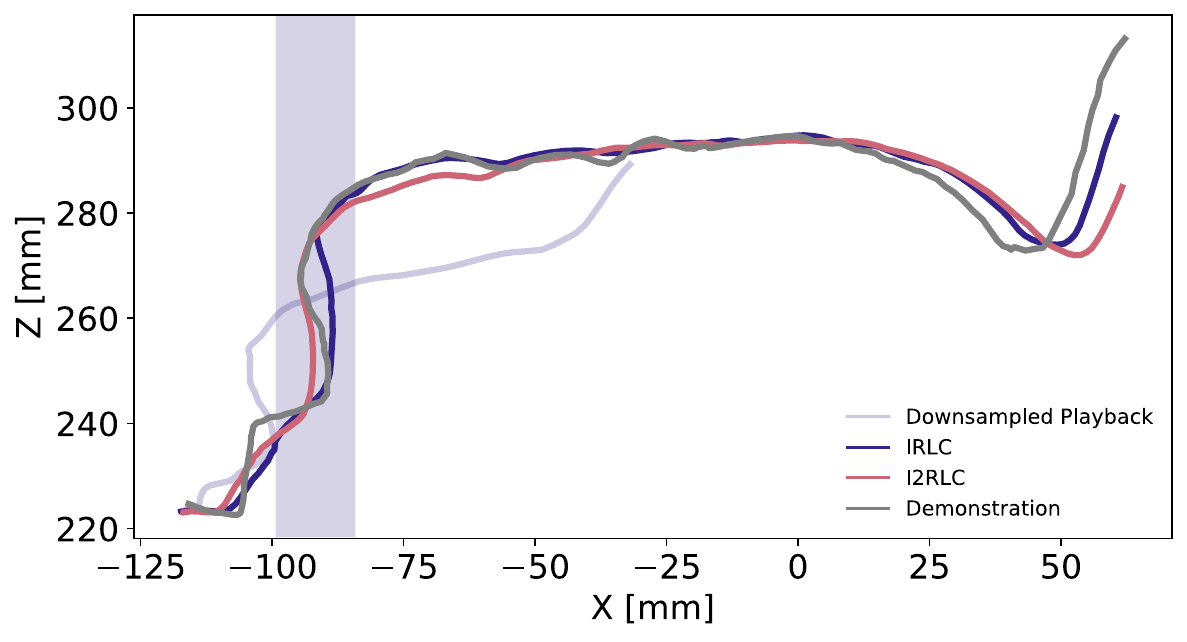}
    \caption{Trajectory comparison for curved surface arc erasing at 10$\times$ speed.}
    \label{fig:trajectory_comparison_curved_surface}
\end{figure}

\begin{table}
    \centering
    \caption{IRLC and I2RLC RMS contact force comparison in peg-in-hole [N] $\downarrow$}
    \label{table:force_comparison}
    \begin{tabular}{c|cccccccc}
        \toprule
        Speed & 3$\times$ & 4$\times$ & 5$\times$ & 6$\times$ & 7$\times$ & 8$\times$ & 9$\times$ & 10$\times$ \\
        \midrule
        IRLC  & 5.69 & 5.73 & 5.80 & 5.87 & 6.05 & 6.44 & 6.88 & 5.95 \\
        I2RLC & \bf{5.59} & \bf{5.60} & \bf{5.56} & \bf{5.78} & \bf{5.50} & \bf{5.95} & \bf{5.59} & \bf{5.54} \\
        \bottomrule
    \end{tabular}
\end{table}

To visualize the effects, \cref{fig:snapshots} presents snapshots from the flat and curved surface erasing tasks executed at the final iteration of 10$\times$ speed. A naively downsampled playback fails to erase most of the blue line (\cref{fig:snapshots}a).  A supplementary video also demonstrates the execution of the naive downsampled replay at the initial iteration. 
IRLC removes the majority of the line but leaves some residual segments (\cref{fig:snapshots}b). By contrast, I2RLC erases the blue line completely (\cref{fig:snapshots}c).

\cref{fig:trajectory_comparison_falt_surface,fig:trajectory_comparison_curved_surface} present 2D plots of the executed trajectories at the final iteration of the 10$\times$ speed stage. While IRLC and I2RLC follow similar paths overall, IRLC tends to achieve smaller errors at larger X values, whereas I2RLC achieves smaller errors at smaller X values. Since errors at smaller X values are larger in magnitude, they dominate the average DTW error, leading I2RLC to achieve a lower overall average DTW error. The larger errors under IRLC within the shaded regions correspond to the residual line segments (\cref{fig:snapshots}b).
\cref{table:force_comparison} shows a root mean square (RMS) contact force along the X, Y, and Z-axes at the final iteration. 
I2RLC exhibits slightly lower RMS contact forces than IRLC, with the difference becoming more pronounced at higher speeds.

In summary, both IRLC and I2RLC effectively refine accelerated trajectories, with I2RLC offering smaller errors in early iterations and better preservation of spatial similarity.

\subsection{Imitation Learning}
\label{sec:IL_exp}

I2RLC is applied to IL across two manipulation tasks: flat-surface arc erasing and peg-in-hole insertion. As IRLC-refined demonstrations for the arc-erasing task exhibited residual line segments, the resulting demonstration quality was deemed insufficient for reliable IL training. Therefore, we trained IL policies using I2RLC-refined demonstrations for the arc-erasing task. For the peg-in-hole task, both IRLC and I2RLC-refined demonstrations successfully completed the insertion, allowing us to compare the downstream IL performance of both methods.

\begin{table}[t]
    \centering
    \caption{
        Erased-area rate with ACT $\uparrow$[\%]. \\      S: Seen (trained) position, U: unseen position.
    }
    \label{table:residual_line_rate}
    \begin{tabular}{c@{\hspace{2pt}}c|ccccccc}
    \toprule
    & Pattern & 1 (U) & 2 (S) & 3 (U) & 4 (S) & 5 (U) & 6 (S) & 7 (U) \\
    \midrule
    \multirow{1}{*}{} & I2RLC & 53±9 & 62±5 & 56±6 & 85±14 & 67±9 & 67±17 & 68±9 \\
    \bottomrule
    \end{tabular}
\end{table}
\begin{figure}[t]
    \centering
    \includegraphics[width=1\linewidth]{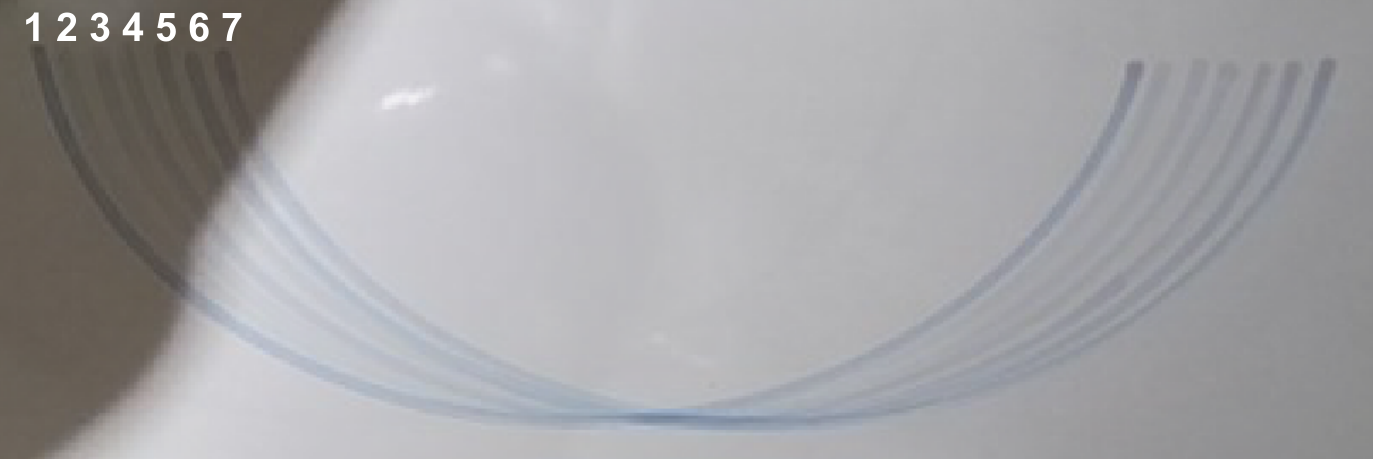}
    \caption{Line patterns of flat surface arc erasing task}
    \label{fig:line_pattern}
\end{figure}

\textbf{Setup:}
For the flat-surface arc erasing task, two demonstrations are collected for each of three arc patterns (six demonstrations total). I2RLC then generates accelerated replays up to 6$\times$
the original speed, incrementally increasing the playback rate with three refinement iterations at each speed stage. Ten rollouts with additive Gaussian noise are executed per demonstration at the final 6$\times$ speed stage, yielding a dataset of 60 episodes.
For the peg-in-hole task, a single demonstration is collected and accelerated up to 10$\times$, following the same refinement procedure. 30 rollouts with Gaussian noise are executed at the final 10$\times$ speed stage, yielding a dataset of 30 episodes.
In both tasks, ACT~\cite{zhao2023learning} is trained for 15,000 steps with a chunk size of 50 using data collected at 50 Hz.

\begin{figure}
    \centering
    \includegraphics[width=1\linewidth]{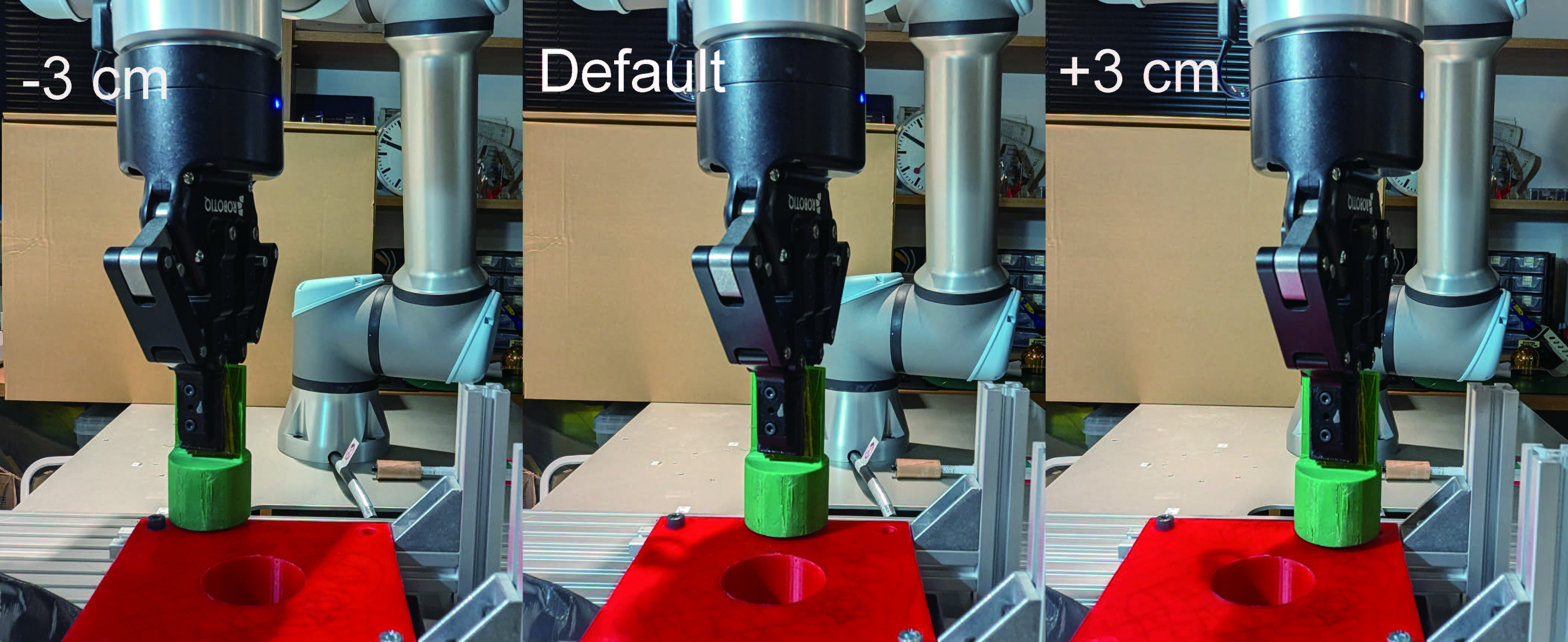}
    \caption{Peg-in-hole initial positions}
    \label{fig:peg}
\end{figure}

At inference time, the trained policies are evaluated on the flat-surface erasing task for both seen and unseen arc positions. The arc position patterns are shown in \cref{fig:line_pattern}.
The erased-area rate is defined as the fraction of the initial blue-line area that is removed. Binary masks are obtained by segmenting the blue line from pre- and post-execution images; the removed area is computed from the mask difference and normalized by the initial mask area.
For the peg-in-hole task, the policy is evaluated at both seen and unseen initial positions, where the unseen positions are displaced by 3 cm from the seen position, as shown in \cref{fig:peg}.
In both tasks, 20 rollouts are performed for each starting position.

\begin{table}[t]
    \centering
    \caption{
        Success rates and 
        contact forces in peg-in-hole tasks with ACT. \\S: Seen (trained) position, U: unseen position.
    }
    \label{table:pegin_success}
    \begin{tabular}{c@{\hspace{-5pt}}c|ccc}
    \toprule
    & Position & $-3$\,cm (U) & $0$\,cm (S) & $+3$\,cm (U) \\
    \midrule
    \multirow{2}{*}{\begin{tabular}{c} Success \\ rate $\uparrow$ \end{tabular}}
      & IRLC & \textbf{100\%} & \textbf{100\%} & \textbf{100\%} \\
      & I2RLC  & \textbf{100\%} & \textbf{100\%} & \textbf{100\%} \\
    \midrule
    \multirow{2}{*}{\begin{tabular}{c} RMS force \\ {[N]} $\downarrow$ \end{tabular}}
      & IRLC  & $7.46 \pm 0.52$ & $6.86 \pm 0.22$ & $6.65 \pm 0.48$ \\
      & I2RLC & $\mathbf{7.15 \pm 0.77}$ & $\mathbf{6.08 \pm 0.52}$ & $\mathbf{6.34 \pm 0.54}$ \\
    \bottomrule
    \end{tabular}
\end{table}

\subsubsection{Results}
\Cref{table:residual_line_rate} reports the erased-area rates measured from images.
The learned policy executed faster than the original demonstration and erased over 65\% of the blue-line segments on average, but it left slightly more residual line area than direct playback.
Two factors may contribute to this gap: 1) the output post-processing (Temporal Ensemble) used in our IL pipeline, which can introduce latency and damp rapid corrections; and 2) occlusions late in the task, when the end effector obscures the target line. 
\cref{table:pegin_success} shows the success rates and mean RMS contact forces with standard deviations over 20 rollouts in the peg-in-hole task. Both I2RLC- and IRLC-trained policies achieve 100\% success rates at seen and unseen initial positions. Furthermore, I2RLC exhibits slightly lower contact forces than IRLC across all positions.

In summary, ACT with faster demonstrations achieved an average erasure of 65\% of the blue-line segments in the arc erasing task and 100\% success rates in the peg-in-hole tasks. 

\section{Conclusion}

We introduce two methods for high-speed replay of demonstrations in contact-rich IL without human intervention. First, IRLC is repurposed as a demonstration acceleration framework, and the results show that it substantially reduces tracking errors relative to naive speed-up playback. Second, I2RLC is proposed as an enhancement of IRLC through incremental speed scheduling and warm-start initialization, yielding improved spatial fidelity and more stable convergence.
Both methods achieve up to 10$\times$ speedup, though IRLC required early termination in the peg-in-hole task at 10$\times$ due to protective stop activation. I2RLC-refined demonstrations enabled ACT policies achieving 100\% success rates in the peg-in-hole task at both seen and unseen positions with lower contact forces than policies trained on IRLC-refined data, and erasing over 65\% of the target line segments in the arc-erasing task. A current limitation is the iteration cost required to achieve large speedups. Future work includes optimizing learning gains and speed-scheduling policies, as well as developing faster IL algorithms using the refined accelerated demonstrations.



\bibliographystyle{IEEEtran}
\bibliography{reference}

\end{document}